\documentclass[12pt]{article}

\usepackage{sbc-template}
\usepackage{graphicx,url}
\usepackage[utf8]{inputenc}
\usepackage[english]{babel}

\usepackage{booktabs}
\usepackage{multicol}
\usepackage{multirow}

\title{The Newsworthiness of Brazilian Distress: A Peak Analysis on Time Series of International Media Attention to Disasters in Brazil}

\author{Brielen Madureira\inst{1,2}, Andreas Niekler\inst{1,3}, Marc Keuschnigg\inst{4,5}, Mariana Madruga de Brito\inst{2}}

\address{LeipzigLab - Climate Discourse, Leipzig University, Germany
\nextinstitute
  Helmholtz Centre for Environmental Research - UFZ, Germany
\nextinstitute
	Computational Humanities, Leipzig University, Leipzig, Germany
\nextinstitute
	Institute of Sociology, Leipzig University, Germany
\nextinstitute
	Institute for Analytical Sociology, Linköping University, Sweden
  \email{brielen.madureira@uni-leipzig.de}
}

\begin{document} 

\maketitle

\begin{abstract}
 Media coverage influences disaster response, yet the drivers of international media attention to local events remain unevenly understood. Brazil offers a compelling case: some of its natural and technological disasters occasionally hit the international headlines. However, systematic analyses of what makes these events be discussed abroad are still missing. Addressing this gap requires representative, validated and country-specific news datasets. This paper presents a peak analysis of 2k news about Brazilian fires and landslides in German newspapers from 2000 to 2024. Using time series segmentation to detect news event peaks, we examine the extent to which they can be temporally aligned with observations in national and global disaster databases.
\end{abstract}

\section{Introduction} \label{sec:intro}

\begin{figure}[ht]
	\centering
	\includegraphics[trim={0cm 5.3cm 0 0},clip,width=\textwidth]{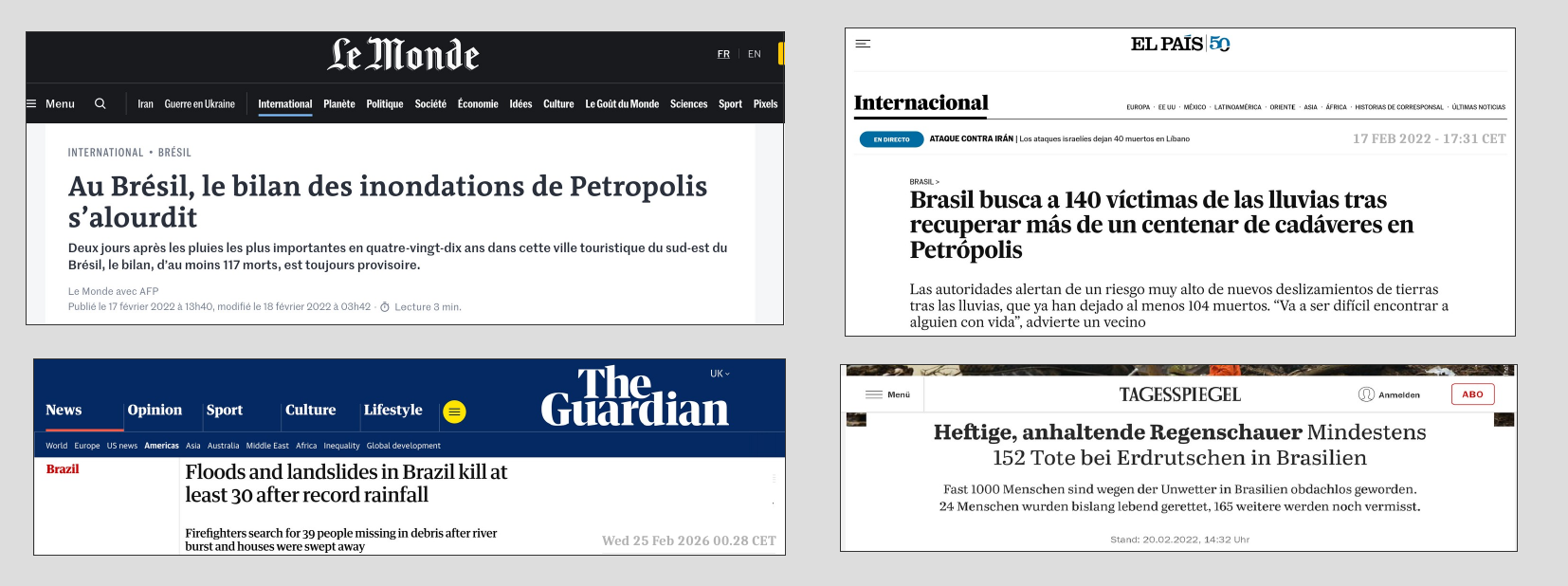}
	\caption{International news headlines about the fatal floods and landslides in Petrópolis (RJ), in February, 2022}
	\label{fig:headlines}
\end{figure}

On February 15, 2022, intense rainfall triggered landslides and floods in Petrópolis-RJ, Brazil, resulting in 231 fatalities and further impacts.\footnote{Source: \url{https://en.wikipedia.org/wiki/2022_Petr\%C3\%B3polis_floods}} This tragedy did not go unnoticed abroad: as we see in Figure \ref{fig:headlines}, various media outlets around the world reported on this event. But not every disaster in Brazil becomes so internationally prominent. The flow of news is shaped by a myriad of factors, from political and economic decisions, to disaster attributes (e.g.~their severity) and aspects inherent to the news publication and dissemination process such as sensationalism, simplification and newsworthiness \cite{stgaard1965,Yan2015}. Social science studies of \textit{collective attention} in the media seek to explain how information emerges in public discourse, remains under discussion and is eventually forgotten \cite{Candia2018,LorenzSpreen2019}. 

News coverage can influence public opinion, political decision-making and allocation of humanitarian aid \cite{Olsen2003,Eisensee2007,Miles2007,Sloggy2021}. Computational methods can aid in unravelling the dynamics between disasters, public discourse and political action. In quantitative news assessments, some measurable characteristics of these dynamics are, for instance, the time needed for an event to appear in the news, how many media outlets cover the topic, the total volume of articles about it, for how long it remains newsworthy, how gradually it fades, and how sporadically it is recalled afterwards. 

As we discuss in Section \ref{sec:lit}, research on media coverage of disasters in Brazil typically focuses on one or a few specific events and analyses them only within the scope of the Brazilian press. To our knowledge, no large scale longitudinal analyses regarding \textit{international} media coverage of multiple disasters in Brazil exists. Addressing this gap requires high quality datasets of international news and suitable natural language processing (NLP) methods to identify and classify relevant documents, geoparse them and extract information on the disaster's framing and impacts. To support reliable downstream conclusions, such datasets must be carefully analysed and \textit{validated}. This includes assessing their representativeness, measuring precision and recall, and examining their temporal alignment with external disaster databases describing the underlying phenomena.

In previous work, we \cite{madureira-2026-geo,madureira-2026-tm} explored methods for constructing and geolocating a dataset of news articles about worldwide disasters and extreme climate events in German-language news. We aimed to enable computational text analyses of climate discourse in the media and to model variations in attention curves across countries. However, this dataset has neither been validated nor analysed at the country-level yet. In this paper, we take Brazil as a case study and present a proof of concept analysis of news event detection via time series segmentation and alignment with two authoritative disaster databases: the global EM-DAT \cite{Delforge2025} and the national S2iD-based Digital Atlas of Disasters in Brazil.\footnote{Available at \url{https://atlasdigital.mdr.gov.br/}} Moreover, we manually verify the capability of the resulting dataset to represent the phenomena of interest. This work lays the foundation for upcoming research on modelling the characteristics (e.g.~number of fatalities) that make a disaster in Brazil newsworthy abroad.

\section{Related Work} \label{sec:lit}

The motivation for this study is threefold: the need for validated datasets in the field of climate impact and adaptation, given the limitations of global disaster databases; open questions about how climate discourse triggers political action; and the research gap regarding attention cycles to Brazilian disasters in international media.

\textbf{Text-based data for socio-environmental research}: With the popularisation of NLP tools and methods, large-scale text corpora are being increasingly used for information extraction and discourse analysis in research fields like climate impact and adaptation, disaster relief and media studies. News databases are a typical source of text data: recent work has examined e.g.~frames and actors related to carbon capture and storage in German newspapers \cite{Otto2022}, how climate change coverage in German media is affected by sociopolitical and extreme weather events \cite{Lochner2024}, the country-level attention to climate-related disasters in British newspapers \cite{Kong2025}, and the social vs.~traditional media coverage of rainstorms in China \cite{Zheng2024}. Text-mining and information extraction techniques are extensively used to derive datasets of socio-economic impacts of extreme climate events, e.g.~droughts \cite{Sodoge2024} and flash floods \cite{HenriqueLimaAlencar2024} as well as multi-hazard analyses \cite{li-etal-2024-using-llms}.

\textbf{Limitations of existing global databases}: Findings that rely on global disaster databases as official compilations of known events are subject to their inherent biases \cite{Gall2009}. \cite{Panwar2019} showed that reported estimates vary considerably between the two most widely used references: the Emergency Events Database (EM-DAT \cite{Delforge2025}) and DesInventar \cite{desinventar}. For instance, EM-DAT's inclusion criteria impose that the disaster must cause at least 10 deaths, affect at least 100 people or result in a call for international assistance or a declaration of a state of emergency. DesInventar currently aggregates inventories of 86 countries, but Brazil is not included. In Brazil, the Integrated Disaster Information System (S2iD) is the official source for national disaster reports from local governments. As not every country discloses reliable databases, other types of (unstructured) data, such as news articles and reports, can provide additional information. They also have the potential to cover events that did not meet the inclusion criteria of global databases.

\textbf{Validation in computational text analysis}: In fields that rely on computational methods to perform content and text analysis, the concept of \textit{validation} is fundamental in operationalising constructs \cite{Grimmer2013,Baden2021,Birkenmaier2023search}. In this context, to validate means to assess ``whether measures actually measure what they are designed to measure'' \cite{BernhardHarrer2025}. Various types of validity have been categorised in the literature \cite{birkenmaier2023,BernhardHarrer2025}. Our study performs \textit{internal validation} of the outputs and \textit{external validation}, using ``external criteria unrelated to the textual data itself'' \cite{Birkenmaier2023search} to assess the plausibility of insights derived from its measures.

\textbf{Media coverage of disasters in Brazil}: Although a large body of work has been conducted in this field, studies mostly investigate single disasters (e.g.~the 2015 Mariana dam collapse \cite{Mouro2018,Prado2020}, the consequences of heavy rainfalls in Santa Catarina in 2008 and in the Serrana highlands of the state of Rio de Janeiro in 2011 \cite{Lahsen2019} and the tragic floods in Rio Grande do Sul in May, 2024 \cite{Silva2025}) or focus on broad environmental topics (e.g.~climate change or crisis \cite{Dayrell2019,loose2023public}). Also, all these studies examined only Brazilian media, disregarding how Brazilian disasters are covered internationally. However, understanding international coverage is critical as it shapes global awareness, foreign policy, and public perception. An exception is \cite{buarque2023}, who analysed international media coverage across seven newspapers of different countries, but also on a single event (the murder of a Brazilian indigenist and a British journalist while working to protect the Amazon Forest). Our work differs in that it examines multiple events over 24 years, as covered by hundreds of news outlets in Germany, and aims to be indicative of how Brazil is generally portrayed to the German audience.

\section{Data, Methods and Measures} \label{sec:method}

Our analysis consists of four parts: (i) descriptive statistics of the data sample; (ii) characterisation of each hazard's count time series based on peak analysis;\footnote{See \url{https://www.mathworks.com/help/signal/ug/peak-analysis.html}} (iii) validation of the identified news events against external databases of disasters; (iv) manual error analysis and hazard-specific findings.

Our dataset was presented in detail by \cite{madureira-2026-tm}, thus only its main characteristics are summarised here. The authors queried the wiso-net\footnote{Available at \url{https://www.wiso-net.de/}} news aggregator database using a curated list of hazard-related German keywords to retrieve documents about extreme climate events worldwide in the period from 2000 to 2024. Relevant articles were classified using topic models. After filtering for relevant documents using their balanced topic model classifier, the collection contains around 130k news articles in German divided into two types of hazards: landslide (e.g.~mass movements, debris flow and related processes) and fire (urban and wildfires). The original paper evaluated performance based on a sample of human annotated data, but detailed validation is lacking. Apart from a pilot geolocation study on an annotated sample, metadata identifying the event's location in all documents has not been released. Thus, for this study, we rely on a heuristic to identify whether the event of interest happened in Brazil: the document must contain the word \textit{Brasilien} (Brazil) and no other country name. The rationale is that a newspaper in Germany is unlikely to refer to a location in Brazil without specifying the country.\footnote{This inevitably leads to some false positives and negatives, but we assume that the sampling retains the needed overall properties. Some exceptions, like the prominence of Rio de Janeiro and the Amazon region as tourist destinations, cannot be ruled out. Texts about  multiple countries must first be properly geolocated (see the Limitations section).} 

We take the daily count of news articles as a proxy for collective attention: the more news articles about a topic in a day, the more attention it attracts. If the same text is published by different news outlets, we count each as a separate observation. We construct a count time series \cite{Davis2021} for each hazard type, i.e.~a time series with non-negative, integer values representing the daily frequency of news articles about events in Brazil from Jan 1, 2000 to Dec 31, 2024. Each time series is segmented into what we call \textit{news events} using Scipy's \texttt{find\_peaks} method.\footnote{Available at \url{https://docs.scipy.org/doc/scipy/reference/generated/scipy.signal.find_peaks.html}} This method identifies peak dates, so that each peak and its neighbouring dates with at least one observation are considered a separate \textit{news event}, surrounded by days without any news. The two main hyperparameters are the minimum peak height, set to 2 in order to reduce single article peaks without much impact in attention, and the minimum distance in days between peaks, set to 7 to represent a one-week interval. We then characterise news events for each hazard using nine measures, listed in Table \ref{tab:measures}.  

\begin{table}[t]
	\centering
	\small
	\caption{Overview of the measures used for the peak analysis}
	\begin{tabular}{ll}
		\toprule
		\textbf{measure} & \textbf{description} \\
		\midrule
		$n$ events & number of identified peaks (news events) \\
		$n$ at peak & daily article count at the event's peak date \\
		total volume & total number of articles during the news event \\
		duration & total number of days in the news event \\
		days since last & number of days since the last news event \\
		days to peak & number of days from the first article's date until the peak date \\
		days to fade & number of days from the peak date to the last article's date \\
		$n$ text types & number of unique text types published during the news event \\
		$n$ outlets & number of media outlets reporting during the news event \\
		\cline{1-2}
		\bottomrule
	\end{tabular}
	\label{tab:measures}
\end{table}

For validation, we temporally align identified news events to known disasters in two external sources, namely the EM-DAT and S2iD databases. We consider a news event (the peak and its neighbouring days) to be aligned with a disaster database entry if the disaster's official onset date falls between 5 and 0 days from the first news event date. A news event can be aligned to more than one disaster event that occurs concurrently. Finally, we manually inspect the content of all detected news events and aligned disaster entries to validate the results.

\section{Analysis}  \label{sec:analysis}

\begin{figure}[t]
	\centering
	\includegraphics[trim={0cm 0cm 0 0},clip,width=\textwidth]{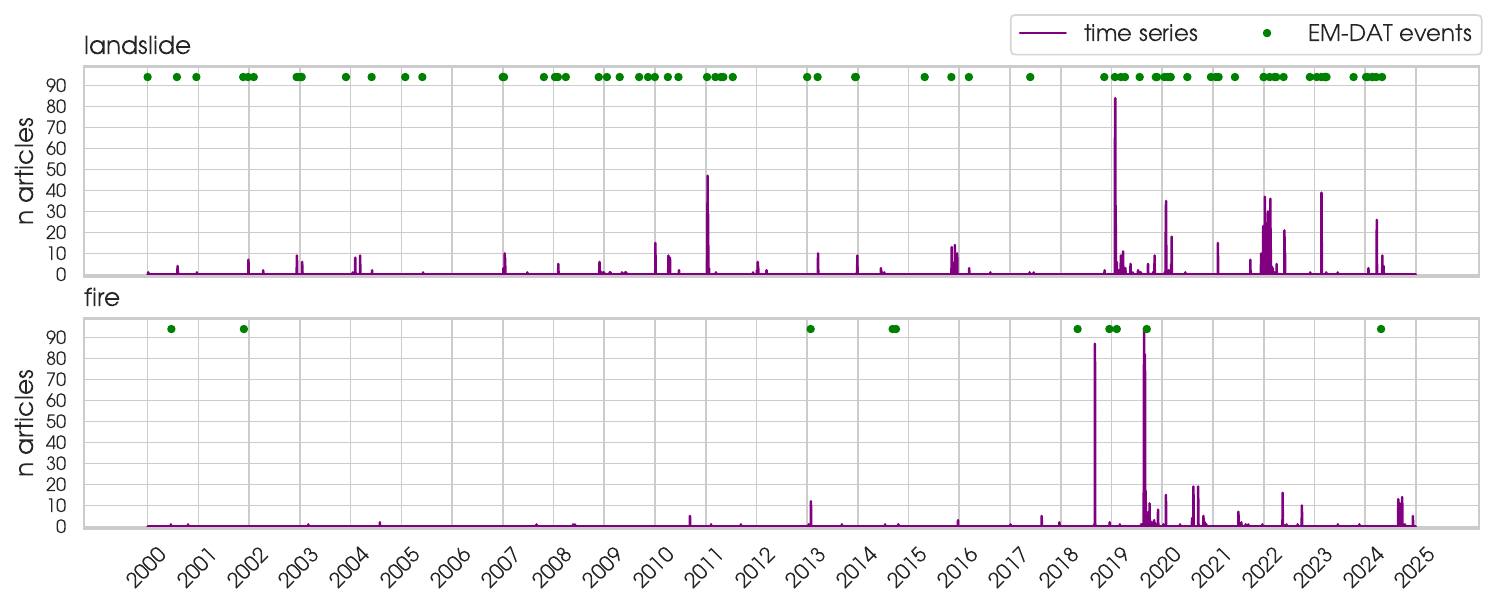}
	\caption{Overview of the time series for each hazard from 2000 to 2004}
	\label{fig:series}
\end{figure}

\paragraph{Descriptive statistics} The news sample representing Brazil has 2,118 documents with, on average, 11.6 (std$=$9.3) sentences and 199.4 (std$=$172.0) tokens. It contains news published by 219 news outlets.\footnote{The top five are \textit{Neue Westfälische}, \textit{Spiegel Online}, \textit{Berliner Zeitung}, \textit{Fränkischer Tag} and \textit{Südkurier} with between 31 and 49 articles each.} Table \ref{tab:data} shows its basic descriptive measures: the number of news articles and unique texts per hazard, the maximum daily count and the number of days with a count larger than 0 out of the total 9,132 days. The vast majority of days (more than 97\% for each hazard) has no observations, resulting in a very sparse signal. This is expected, as disasters are, by their nature, extreme. The two rightmost columns show the mean count and standard deviation when only active days are considered (since, given the sparseness, the means are below 0.14 if all days in the period are considered). Landslide news occur more often and have a larger total amount, but the daily mean and maximum are slightly higher for fires. Figure \ref{fig:series} illustrates the series' shape and peaks and the dates of EM-DAT events. News about fires peak rarely until around 2013. The highest peaks occur in 2018 and 2019, followed by several others. Landslide news peak more often throughout the period with a prominent one at the beginning of 2019. The actual disasters are identified in the final discussion of this section. The news frequency increases over time: although this suggests more disasters, it also mirrors the inherent temporal bias toward recent years in the news database and possibly an increase in attention allocated to Brazil in German news (e.g.~following the World Cup of 2014).

\begin{table}[h]
	\centering
	\small
	\caption{Overview of the data sample with news about disasters in Brazil}
	\begin{tabular}{lrrr|rrr}
		\toprule
		& $n$ articles & $n$ text types & daily max & $n$ active days & daily mean & daily std \\
		\midrule
		landslide & 1,276 & 602 & 84 & 195 & 6.5 & 10.9 \\
		fire & 842 & 258 & 94 & 112 & 7.5 & 17.1 \\
		\cline{1-6}
		\bottomrule
	\end{tabular}
	\label{tab:data}
\end{table}

\begin{figure}[t]
	\centering
	\includegraphics[trim={0cm 0cm 0 0},clip,width=\textwidth]{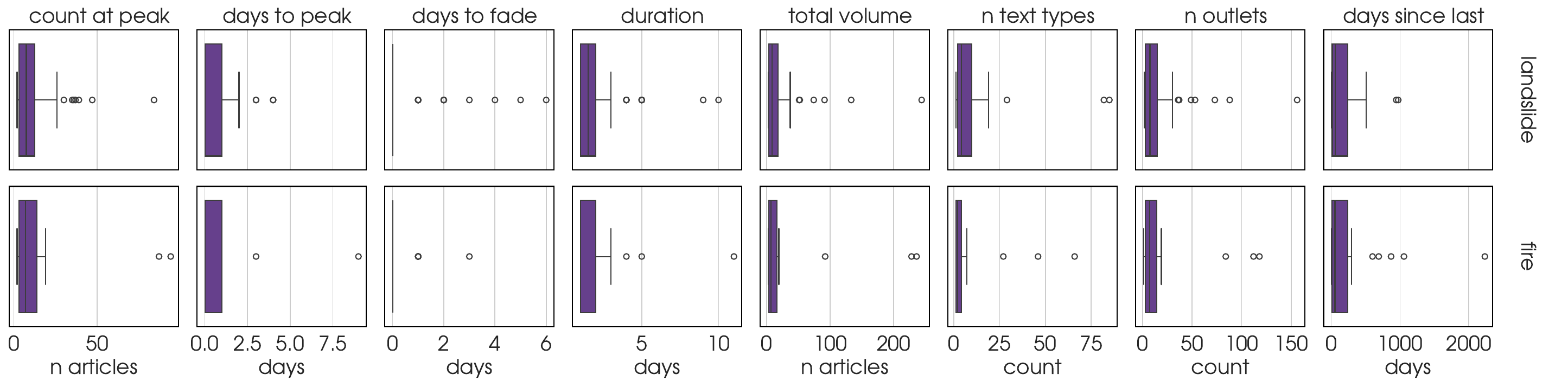}
	\caption{Box plots for all measures and hazard types}
	\label{fig:measures}
\end{figure}

\paragraph{Peak characterization} Our method identified 58 peaks for landslide and 30 for fire, each representing a news event. Figure \ref{fig:measures} shows the box plots of all computed measures, illustrating their distributions. The peaks for fire and landslide have a similar median height (7 and 7.5, respectively). The median of days to peak and to fade is 0 for both hazards. Besides, days to fade shows no spread, except for a few outliers. This indicates that the time series consist of many one or two-day bursts instead of curves with a gradual onset and offset. Indeed, the median coverage duration is 1-1.5 days, and the third quartiles reach 2 days. The median number of media outlets covering fires (7) and of the overall number of articles for an event (7) is similar to landslide (7.5 and 8.5, respectively). In terms of text types, the median for landslide (4) is twice that for fire (2). Finally, the median number of days between two news events is similar for fire (55 days) and  landslide (59 days). We can observe a few outliers that deviate considerably from the typical empirical observations for both hazards, with peaks that are higher and news events with a larger volume, duration, text diversity and outlet diversity. 

\paragraph{Alignment with disaster databases} EM-DAT contains 11 entries about fires and 79 about landslides in Brazil in the same time period, either as the main event or an associated type, including both natural and technological disasters. S2iD has 615 fire and 199 landslide notifications categorised as ``recognised'' (i.e.~excluding those just ``registered'') natural disasters. Out of the 88 news events, 27 (3 for fire and 24 for landslide) and 21 (7 for fire and 14 for landslide) can be temporally aligned to EM-DAT and S2iD entries, respectively. It means that around 28\% of the identified news events can be automatically aligned to a known disaster but many entries in the databases do not temporally align to any news event. The news stream also clusters into events seemingly not included in these disaster databases. This indicates the potential to use news as a complementary data source for disaster identification. However, we must first investigate this misalignment in more detail, as temporal alignment does not always imply content alignment.

\paragraph{Error analysis} To understand what kinds of news events are detected and which accurately align with known disasters, we manually inspect all peaks and news clusters. 

\vspace{0.5cm}

\begin{itemize}
	\item \textbf{Fire}: Among the 30 detected fire-related news events, only one is a false positive (i.e.~the news mention forest overexploitation in Brazil but in the context of a fire in Germany). Four news events are well-known urban fires: the fire in the night club in Santa Maria in 2013, in the Portuguese Language Museum in São Paulo in 2015, in the National Museum in Rio de Janeiro in 2018 (the second highest peak with 87 news in a day) and in a neighbourhood in Manaus in 2018. The first and the last correctly align with EM-DAT entries. At least 21 news events, including two sequential ones with the highest total volumes of more than 200 news, focus on wildfires in the Amazon region, especially since mid 2019. Some of them also refer to the Pantanal and the Cerrado regions. Many of these news are not about a specific disaster, but about the aggregated incidence or consequences of wildfires over a period. One EM-DAT alignment is wrong. Aligned S2iD entries are ambiguous: news usually discuss Amazon fires broadly while S2iD refers to specific cities that may or not be the target of the news; besides, Amazon wildfires often resurface as news events, sometimes only referring to past or aggregated events, and their dates coincide with other concomitant S2iD entries. Alignment also fails because the onset date of wildfire entries cannot always be precise.
	
	\vspace{0.5cm}
	
	\item \textbf{Landslide}: All 58 news events are true positives regarding landslides. Several news events refer to the consequences of rainfall, very often in Rio de Janeiro. Three notable news events in terms of total volume are the floods in the Serrana region of Rio de Janeiro in Jan, 2011, in Petrópolis in Feb, 2022 and in the Southeast states in Jan, 2020. A news event unrelated to climate-driven processes refers to an accident during the construction of a metro station in São Paulo in 2007. Four peaks occur during or right after the Mariana dam collapse in 2015. The industrial disaster in Brumadinho in 2019 is a recurring topic: after the first peak with the highest count (84 news) and total volume (244 news) during the actual event, at least 13 news events refer to it again, bringing this disaster back to the collective attention. The Capitólio rockfall in 2022 also attracted considerable attention. There is a bias towards events in the Southeast states: at least 47 news events refer to disasters in this region. While only 2 of the 24 aligned EM-DAT events are incorrect, many cases are ambiguous and 4 are incorrect for S2iD; only 4 of the 14 aligned are clear matches. Some news events should genuinely not align temporally with database entries because they refer back to prior events, such as the Brumadinho disaster. Landslides associated to other hazards are also missed if they are not registered as a main event in S2iD. 
\end{itemize}  

\paragraph{Discussion} Our analysis provides evidence that the bottom-up dataset creation approach resulted in samples with high precision for these two hazards. On the one hand, a portion of the media peaks have been successfully validated against EM-DAT entries. On the other hand, the news also provide information on a few disasters not yet included in existing datasets. Still, the dataset has some drawbacks. First, recall is unknown: it is possible that some news events were excluded by the data selection and peak detection process. Some news events mix different disasters. The single-article peaks excluded from the analysis would represent valid disasters: their properties should be further investigated as a type of disaster that does appear in German media but attracts less attention. Despite the original work's attempt to detect only extreme climate events, other related disasters are present in the dataset, as it is hard to automate the distinction between natural and technological events using NLP methods. While most aligned EM-DAT events correctly match our detected news events, S2iD matches are often misaligned or ambiguous. Even though we excluded events that were registered by local governments but not recognised by the federal agency, the database still contains many local events that may not have been impactful enough to become newsworthy abroad but co-occur with peaks about previous events like Brumadinho. Besides, because the Amazon forest extends across other states, it is hard to pinpoint the exact region to which news about the Amazon refers. For S2iD, temporal alignment is not enough. Refined methods to avoid misalignments are e.g.~a reverse top-down news selection approach using the disaster metadata to create targeted queries for the news database and matching location toponyms for more certainty (although misspellings in database entries and news in foreign languages can be an issue).

\section{Conclusion}\label{sec:conclusion}

News about disasters in Brazil influence how the country is perceived abroad. We presented a peak analysis of the time series of German news articles about two hazard types occurring in Brazil. The main findings are: the media attention time series are sparse and contain many short (one to two-day) spikes; a few outlier news events attract much more attention during more days, with some disasters resurfacing repeatedly, such as the Amazon wildfires and the Brumadinho dam collapse; and news articles and existing disaster databases can complement each other. Although the measures for both hazards are similar, landslides are typically reported as specific events whereas wildfires are aggregated. 

This analysis prepares the way for a wide variety of further research. A way forward is to expand the investigation to other types of hazards, using methods that properly handle their higher rate of unrelated news, and assess how robust the results are to other peak-finding hyperparameters. We also envision multi-country comparative studies using economic, sociological, cultural and geographical measures for modelling the differences in the media attention they receive. Moreover, once news events and database entries are aligned, the known disaster metadata can be used to model the time series behaviour and investigate which disaster characteristics cause an event to be captured by the international media and to trigger waves of collective attention. 

\section*{Limitations}

This is a preliminary and exploratory analysis of the country-level data. We used a simple heuristic to identify a sub-sample of documents about Brazil. For improving recall, the texts should be properly geolocated, which is ongoing work. Further investigation is also needed regarding the effects of using other hyperparameters in the peak finding method.

\bibliographystyle{sbc}
\bibliography{sbc-template}

@article{stgaard1965,
	title = {Factors Influencing the Flow of News},
	volume = {2},
	ISSN = {1460-3578},
	url = {http://dx.doi.org/10.1177/002234336500200103},
	DOI = {10.1177/002234336500200103},
	number = {1},
	journal = {Journal of Peace Research},
	publisher = {Oxford University Press (OUP)},
	author = {\"{O}stgaard,  Einar},
	year = {1965},
	month = mar,
	pages = {39–63}
}

@article{Candia2018,
	title = {The universal decay of collective memory and attention},
	volume = {3},
	ISSN = {2397-3374},
	url = {http://dx.doi.org/10.1038/s41562-018-0474-5},
	DOI = {10.1038/s41562-018-0474-5},
	number = {1},
	journal = {Nature Human Behaviour},
	publisher = {Springer Science and Business Media LLC},
	author = {Candia,  Cristian and Jara-Figueroa,  C. and Rodriguez-Sickert,  Carlos and Barabási,  Albert-László and Hidalgo,  César A.},
	year = {2018},
	month = dec,
	pages = {82–91}
}

@article{LorenzSpreen2019,
	title = {Accelerating dynamics of collective attention},
	volume = {10},
	ISSN = {2041-1723},
	url = {http://dx.doi.org/10.1038/s41467-019-09311-w},
	DOI = {10.1038/s41467-019-09311-w},
	number = {1},
	journal = {Nature Communications},
	publisher = {Springer Science and Business Media LLC},
	author = {Lorenz-Spreen,  Philipp and Mønsted,  Bjarke Mørch and H\"{o}vel,  Philipp and Lehmann,  Sune},
	year = {2019},
	month = apr 
}

@article{Olsen2003,
	title = {Humanitarian Crises: What Determines the Level of Emergency Assistance? Media Coverage,  Donor Interests and the Aid Business},
	volume = {27},
	ISSN = {1467-7717},
	url = {http://dx.doi.org/10.1111/1467-7717.00223},
	DOI = {10.1111/1467-7717.00223},
	number = {2},
	journal = {Disasters},
	publisher = {Wiley},
	author = {Olsen,  Gorm Rye and Carstensen,  Nils and Høyen,  Kristian},
	year = {2003},
	month = may,
	pages = {109–126}
}

@article{Miles2007,
	title = {The role of news media in natural disaster risk and recovery},
	volume = {63},
	ISSN = {0921-8009},
	url = {http://dx.doi.org/10.1016/j.ecolecon.2006.08.007},
	DOI = {10.1016/j.ecolecon.2006.08.007},
	number = {2–3},
	journal = {Ecological Economics},
	publisher = {Elsevier BV},
	author = {Miles,  Brian and Morse,  Stephanie},
	year = {2007},
	month = aug,
	pages = {365–373}
}

@article{Sloggy2021,
	title = {Changing opinions on a changing climate: the effects of natural disasters on public perceptions of climate change},
	volume = {168},
	ISSN = {1573-1480},
	url = {http://dx.doi.org/10.1007/s10584-021-03242-6},
	DOI = {10.1007/s10584-021-03242-6},
	number = {3–4},
	journal = {Climatic Change},
	publisher = {Springer Science and Business Media LLC},
	author = {Sloggy,  Matthew R. and Suter,  Jordan F. and Rad,  Mani Rouhi and Manning,  Dale T. and Goemans,  Chris},
	year = {2021},
	month = oct 
}

@article{Dayrell2019,
	title = {Discourses around climate change in Brazilian newspapers: 2003–2013},
	volume = {13},
	ISSN = {1750-4821},
	url = {http://dx.doi.org/10.1177/1750481318817620},
	DOI = {10.1177/1750481318817620},
	number = {2},
	journal = {Discourse \& Communication},
	publisher = {SAGE Publications},
	author = {Dayrell,  Carmen},
	year = {2019},
	month = jan,
	pages = {149–171}
}

@incollection{loose2023public,
	title={Public Communication and Perceptions of Climate Change in Brazil: Eloisa Beling Loose and Anabela Carvalho},
	author={Loose, Eloisa Beling and Carvalho, Anabela},
	booktitle={Climate, science and society},
	pages={58--65},
	year={2023},
	publisher={Routledge},
	url={https://www.taylorfrancis.com/chapters/oa-edit/10.4324/9781003409748-10/public-communication-perceptions-climate-change-brazil-eloisa-beling-loose-anabela-carvalho}
}

@article{Silva2025,
	title = {Communication in Disaster—The Contribution of the Press to Highlighting Vulnerabilities: The Case of Rio Grande Do Sul State, Brazil},
	volume = {14},
	ISSN = {2076-0760},
	url = {http://dx.doi.org/10.3390/socsci14070409},
	DOI = {10.3390/socsci14070409},
	number = {7},
	journal = {Social Sciences},
	publisher = {MDPI AG},
	author = {Silva,  Fernando Pereira and de Moraes,  Osvaldo Luiz Leal and Marques Alves,  Rita de Cassia and Barbosa,  Marcia Cristina and Marengo,  José Antonio},
	year = {2025},
	month = jun,
	pages = {409}
}

@article{Lahsen2019,
	title = {When climate change is not blamed: the politics of disaster attribution in international perspective},
	volume = {158},
	ISSN = {1573-1480},
	url = {http://dx.doi.org/10.1007/s10584-019-02642-z},
	DOI = {10.1007/s10584-019-02642-z},
	number = {2},
	journal = {Climatic Change},
	publisher = {Springer Science and Business Media LLC},
	author = {Lahsen,  Myanna and Couto,  Gabriela de Azevedo and Lorenzoni,  Irene},
	year = {2019},
	month = dec,
	pages = {213–233}
}

@inbook{Prado2020,
	title = {Brazilian Local and National News Coverage of the Samarco Disaster: A Disaster for the Community,  the Corporation or the Environment?},
	ISBN = {9783030337124},
	url = {http://dx.doi.org/10.1007/978-3-030-33712-4_2},
	DOI = {10.1007/978-3-030-33712-4_2},
	booktitle = {Media,  Journalism and Disaster Communities},
	publisher = {Springer International Publishing},
	author = {Prado,  Paola and Pinto,  Juliet},
	year = {2020},
	pages = {19–33}
}

@inbook{Mouro2018,
	title = {Environmental Journalism in Brazil: History,  Characteristics,  and Framing of Disasters},
	ISBN = {9783319705095},
	ISSN = {2634-646X},
	url = {http://dx.doi.org/10.1007/978-3-319-70509-5_4},
	DOI = {10.1007/978-3-319-70509-5_4},
	booktitle = {News Media Coverage of Environmental Challenges in Latin America and the Caribbean},
	publisher = {Springer International Publishing},
	author = {Mourão,  Rachel R. and Sturm,  Heloisa Aruth},
	year = {2018},
	pages = {67–90}
}

@article{buarque2023,
	title = {Fire and Death in the Amazon: The reversal of the forest’s role in the international status of Brazil},
	volume = {12},
	ISSN = {2245-4373},
	url = {http://dx.doi.org/10.25160/bjbs.v12i1.134715},
	DOI = {10.25160/bjbs.v12i1.134715},
	number = {1},
	journal = {Brasiliana: Journal for Brazilian Studies},
	publisher = {Brasiliana - Journal for Brazilian Studies},
	author = {Buarque,  Daniel and Gondim Mariutti,  Fabiana},
	year = {2023},
	month = oct 
}

@article{Sodoge2024,
	title = {Text mining uncovers the unique dynamics of socio-economic impacts of the 2018–2022 multi-year drought in Germany},
	volume = {24},
	ISSN = {1684-9981},
	url = {http://dx.doi.org/10.5194/nhess-24-1757-2024},
	DOI = {10.5194/nhess-24-1757-2024},
	number = {5},
	journal = {Natural Hazards and Earth System Sciences},
	publisher = {Copernicus GmbH},
	author = {Sodoge,  Jan and Kuhlicke,  Christian and Mahecha,  Miguel D. and de Brito,  Mariana Madruga},
	year = {2024},
	month = may,
	pages = {1757–1777}
}

@article{HenriqueLimaAlencar2024,
	title = {Flash droughts and their impacts—using newspaper articles to assess the perceived consequences of rapidly emerging droughts},
	volume = {19},
	ISSN = {1748-9326},
	url = {http://dx.doi.org/10.1088/1748-9326/ad58fa},
	DOI = {10.1088/1748-9326/ad58fa},
	number = {7},
	journal = {Environmental Research Letters},
	publisher = {IOP Publishing},
	author = {Alencar,  Pedro Henrique Lima and Sodoge,  Jan and Nora Paton,  Eva and de Brito,  Mariana Madruga},
	year = {2024},
	month = jun,
	pages = {074048}
}

@article{Eisensee2007,
	title = {News Droughts,  News Floods,  and U. S. Disaster Relief},
	volume = {122},
	ISSN = {1531-4650},
	url = {http://dx.doi.org/10.1162/qjec.122.2.693},
	DOI = {10.1162/qjec.122.2.693},
	number = {2},
	journal = {The Quarterly Journal of Economics},
	publisher = {Oxford University Press (OUP)},
	author = {Eisensee,  T. and Stromberg,  D.},
	year = {2007},
	month = may,
	pages = {693–728}
}

@article{Gall2009,
	title = {When Do Losses Count?: Six Fallacies of Natural Hazards Loss Data},
	volume = {90},
	ISSN = {1520-0477},
	url = {http://dx.doi.org/10.1175/2008BAMS2721.1},
	DOI = {10.1175/2008bams2721.1},
	number = {6},
	journal = {Bulletin of the American Meteorological Society},
	publisher = {American Meteorological Society},
	author = {Gall,  Melanie and Borden,  Kevin A. and Cutter,  Susan L.},
	year = {2009},
	month = jun,
	pages = {799–810}
}

@article{Delforge2025,
	title = {EM-DAT: the Emergency Events Database},
	volume = {124},
	ISSN = {2212-4209},
	url = {http://dx.doi.org/10.1016/j.ijdrr.2025.105509},
	DOI = {10.1016/j.ijdrr.2025.105509},
	journal = {International Journal of Disaster Risk Reduction},
	publisher = {Elsevier BV},
	author = {Delforge,  Damien and Wathelet,  Valentin and Below,  Regina and Sofia,  Cinzia Lanfredi and Tonnelier,  Margo and van Loenhout,  Joris A.F. and Speybroeck,  Niko},
	year = {2025},
	month = jun,
	pages = {105509}
}

@article{Yan2015,
	title = {The Sky Is Falling: Predictors of News Coverage of Natural Disasters Worldwide},
	volume = {45},
	ISSN = {1552-3810},
	url = {http://dx.doi.org/10.1177/0093650215573861},
	DOI = {10.1177/0093650215573861},
	number = {6},
	journal = {Communication Research},
	publisher = {SAGE Publications},
	author = {Yan,  Yan and Bissell,  Kim},
	year = {2015},
	month = feb,
	pages = {862–886}
}

@article{Kong2025,
	title = {Analyzing Geographic Bias of Newspaper Articles Reporting Global Climate Disasters},
	volume = {116},
	ISSN = {2469-4460},
	url = {http://dx.doi.org/10.1080/24694452.2025.2564220},
	DOI = {10.1080/24694452.2025.2564220},
	number = {2},
	journal = {Annals of the American Association of Geographers},
	publisher = {Informa UK Limited},
	author = {Kong,  Inhye and Purves,  Ross S.},
	year = {2025},
	month = oct,
	pages = {270–288}
}

@article{Zheng2024,
	title = {Reducing Social Media Attention Inequality in Disasters: The Role of Official Media During Rainstorm Disasters in China},
	volume = {15},
	ISSN = {2192-6395},
	url = {http://dx.doi.org/10.1007/s13753-024-00562-w},
	DOI = {10.1007/s13753-024-00562-w},
	number = {3},
	journal = {International Journal of Disaster Risk Science},
	publisher = {Springer Science and Business Media LLC},
	author = {Zheng,  Longfei and Chen,  Lei and Long,  Fenjie and Liu,  Jianing and Li,  Lei},
	year = {2024},
	month = jun,
	pages = {388–403}
}

@article{Lochner2024,
	title = {Climate summits and protests have a strong impact on climate change media coverage in Germany},
	volume = {5},
	ISSN = {2662-4435},
	url = {http://dx.doi.org/10.1038/s43247-024-01434-3},
	DOI = {10.1038/s43247-024-01434-3},
	number = {1},
	journal = {Communications Earth \& Environment},
	publisher = {Springer Science and Business Media LLC},
	author = {Lochner,  Jakob H. and Stechemesser,  Annika and Wenz,  Leonie},
	year = {2024},
	month = may 
}

@inproceedings{li-etal-2024-using-llms,
	title = "Using {LLM}s to Build a Database of Climate Extreme Impacts",
	author = {Li, Ni  and
	Zahra, Shorouq  and
	Brito, Mariana  and
	Flynn, Clare  and
	G{\"o}rnerup, Olof  and
	Worou, Koffi  and
	Kurfali, Murathan  and
	Meng, Chanjuan  and
	Thiery, Wim  and
	Zscheischler, Jakob  and
	Messori, Gabriele  and
	Nivre, Joakim},
	booktitle = "Proceedings of the 1st Workshop on Natural Language Processing Meets Climate Change (ClimateNLP 2024)",
	month = aug,
	year = "2024",
	address = "Bangkok, Thailand",
	publisher = "Association for Computational Linguistics",
	url = "https://aclanthology.org/2024.climatenlp-1.7/",
	doi = "10.18653/v1/2024.climatenlp-1.7",
	pages = "93--110"
}

@article{Otto2022,
	title = {Fixed Amidst Change: 20 Years of Media Coverage on Carbon Capture and Storage in Germany},
	volume = {14},
	ISSN = {2071-1050},
	url = {http://dx.doi.org/10.3390/su14127342},
	DOI = {10.3390/su14127342},
	number = {12},
	journal = {Sustainability},
	publisher = {MDPI AG},
	author = {Otto,  Danny and Pfeiffer,  Maria and de Brito,  Mariana Madruga and Gross,  Matthias},
	year = {2022},
	month = jun,
	pages = {7342}
}

@article{Panwar2019,
	title = {Disaster Damage Records of EM-DAT and DesInventar: A Systematic Comparison},
	volume = {4},
	ISSN = {2511-1299},
	url = {http://dx.doi.org/10.1007/s41885-019-00052-0},
	DOI = {10.1007/s41885-019-00052-0},
	number = {2},
	journal = {Economics of Disasters and Climate Change},
	publisher = {Springer Science and Business Media LLC},
	author = {Panwar,  Vikrant and Sen,  Subir},
	year = {2019},
	month = dec,
	pages = {295–317}
}

@misc{desinventar,
	title={DesInventar Sendai: Sendai Framework for Disaster Risk Reduction},
	author={UNDRR},
	url={https://www.desinventar.net/index.html},
	year={2015}
}

@article{BernhardHarrer2025,
	title = {Beyond standardization: a comprehensive review of topic modeling validation methods for computational social science research},
	ISSN = {2049-8489},
	url = {http://dx.doi.org/10.1017/psrm.2025.10008},
	DOI = {10.1017/psrm.2025.10008},
	journal = {Political Science Research and Methods},
	publisher = {Cambridge University Press (CUP)},
	author = {Bernhard-Harrer,  Jana and Ashour,  Randa and Eberl,  Jakob-Moritz and Tolochko,  Petro and Boomgaarden,  Hajo},
	year = {2025},
	month = jun,
	pages = {1–19}
}

@article{Birkenmaier2023search,
	title = {The Search for Solid Ground in Text as Data: A Systematic Review of Validation Practices and Practical Recommendations for Validation},
	volume = {18},
	ISSN = {1931-2466},
	url = {http://dx.doi.org/10.1080/19312458.2023.2285765},
	DOI = {10.1080/19312458.2023.2285765},
	number = {3},
	journal = {Communication Methods and Measures},
	publisher = {Informa UK Limited},
	author = {Birkenmaier,  Lukas and Lechner,  Clemens M. and Wagner,  Claudia},
	year = {2023},
	month = nov,
	pages = {249–277}
}

@article{Baden2021,
	title = {Three Gaps in Computational Text Analysis Methods for Social Sciences: A Research Agenda},
	volume = {16},
	ISSN = {1931-2466},
	url = {http://dx.doi.org/10.1080/19312458.2021.2015574},
	DOI = {10.1080/19312458.2021.2015574},
	number = {1},
	journal = {Communication Methods and Measures},
	publisher = {Informa UK Limited},
	author = {Baden,  Christian and Pipal,  Christian and Schoonvelde,  Martijn and van der Velden,  Mariken A. C. G},
	year = {2021},
	month = dec,
	pages = {1–18}
}

@article{Grimmer2013,
	title = {Text as Data: The Promise and Pitfalls of Automatic Content Analysis Methods for Political Texts},
	volume = {21},
	ISSN = {1476-4989},
	url = {http://dx.doi.org/10.1093/pan/mps028},
	DOI = {10.1093/pan/mps028},
	number = {3},
	journal = {Political Analysis},
	publisher = {Cambridge University Press (CUP)},
	author = {Grimmer,  Justin and Stewart,  Brandon M.},
	year = {2013},
	pages = {267–297}
}

@misc{birkenmaier2023,
	doi = {10.48550/arxiv.2307.02863},
	url = {https://arxiv.org/abs/2307.02863},
	author = {Birkenmaier,  Lukas and Wagner,  Claudia and Lechner,  Clemens},
	title = {ValiText -- a unified validation framework for computational text-based measures of social constructs},
	publisher = {arXiv},
	year = {2023},
	copyright = {Creative Commons Attribution Non Commercial Share Alike 4.0 International}
}

@article{Davis2021,
	title = {Count Time Series: A Methodological Review},
	volume = {116},
	ISSN = {1537-274X},
	url = {http://dx.doi.org/10.1080/01621459.2021.1904957},
	DOI = {10.1080/01621459.2021.1904957},
	number = {535},
	journal = {Journal of the American Statistical Association},
	publisher = {Informa UK Limited},
	author = {Davis,  Richard A. and Fokianos,  Konstantinos and Holan,  Scott H. and Joe,  Harry and Livsey,  James and Lund,  Robert and Pipiras,  Vladas and Ravishanker,  Nalini},
	year = {2021},
	month = may,
	pages = {1533–1547}
}

@inproceedings{madureira-2026-tm,
	title = "Retrieving Floods without Floodlights: Topic Models as Binary Classifiers for Extreme Climate Events in German News",
	author = "Madureira, Brielen  and
	Brito, Mariana Madruga de and
	Niekler, Andreas",
	booktitle = "Proceedings of the 2nd Workshop on Ecology, Environment, and Natural Language Processing (NLP4Ecology2026)",
	month = may,
	year = "2026",
	doi = "https://doi.org/10.48550/arXiv.2605.03450",
	url = "https://doi.org/10.48550/arXiv.2605.03450"
}

@inproceedings{madureira-2026-geo,
	title = "Geolocating News about Extreme Climate Events: A Comparative Analysis of Off-the-Shelf Tools for Toponym Identification in German",
	author = "Madureira, Brielen  and
	Brito, Mariana Madruga de and
	Niekler, Andreas",
	booktitle = "Proceedings of the Fourth International Workshop on Geographic Information Extraction from Texts (GeoExT 2026) co-located with 48th European Conference on Information Retrieval (ECIR 2026)",
	month = april,
	year = "2026",
	url = "https://ceur-ws.org/Vol-4201/paper1.pdf"
}

\end{document}